\documentclass[runningheads]{llncs}

\usepackage[T1]{fontenc}
\usepackage{graphicx,verbatim}
\usepackage{amsmath}
\usepackage{amssymb}
\usepackage{url}
\usepackage{xcolor}
\definecolor{linkmagenta}{RGB}{236,0,140}

\begin{document}

\title{
HeartVolMesh: Cardiac Volumetric Mesh Reconstruction via Covariance-Guided Graph Deformation
}

\author{Fengming Lin\inst{1,2,3} \and
Arezoo Zakeri\inst{1,2,4} \and
Haoran Dou\inst{1,2,3} \and
Zherui Zhou\inst{1,2} \and
Shaokun Lan\inst{1,2,3} \and
Jinming Duan\inst{1,2,4} \and
Alejandro Frangi\inst{1,2,3,4,5}\thanks{Corresponding author.}}
\authorrunning{F. Lin et al.}
\institute{Centre for Computational Imaging and Modelling in Medicine (CIMIM), University of Manchester, Manchester, UK \and
Christabel Pankhurst Institute, University of Manchester, Manchester, UK \and
Department of Computer Science, University of Manchester, Manchester, UK \and
Division of Informatics, Imaging \& Data Sciences, University of Manchester, Manchester, UK \and
NIHR Manchester Biomedical Research Centre, Manchester Academic Health Sciences Centre, University of Manchester, Manchester, UK\\
    \email{alejandro.frangi@manchester.ac.uk}\\
    {\url{https://fmlinks.github.io/projects/heartvolmesh/}}\\
    {\url{https://github.com/ccmim/HeartVolMesh}}}
  
\maketitle              
\begin{abstract}

Accurate patient-specific tetrahedral cardiac meshes are essential for in-silico trials, yet common segmentation-then-modelling pipelines can blur thin-wall anatomy and offer limited cross-case correspondence. We propose HeartVolMesh, which lifts each template vertex to an anisotropic Gaussian kernel and uses a 3D CNN-GNN to predict per-vertex displacements and Cholesky-parameterized covariances from volumetric images. Training is guided by a covariance-aware negative log-likelihood loss with lightweight mesh regularization. For volumetric meshing, we warp a fixed tetrahedral template to the reconstructed surface via staged alignment, non-rigid registration, and deformation propagation, preserving connectivity and correspondence by construction, with resolution controlled by template density. Experiments show consistent gains over deformation-based baselines in surface mesh accuracy and volumetric mesh fidelity.


\end{abstract}

\section{Introduction}

In-silico trials~\cite{ref_viceconti2016,ref_rodero2023,ref_pathmanathan2024} increasingly evaluate cardiovascular devices and treatments using cardiac digital twins~\cite{ref_sel2024}, reducing reliance on costly clinical studies~\cite{ref_samei2025,ref_pathmanathan2024}. For FEM/CFD solvers, such twins require not only accurate anatomy but also directly usable 3D volumetric, typically tetrahedral, meshes~\cite{ref_fedele2021,ref_simvascular2016}. Automatically reconstructing topology-consistent, detail-faithful, high-quality cardiac volumetric meshes~\cite{ref_survey_pixels2polygons2025} from 3D medical images such as CTA is therefore central to scalable digital twins and in-silico trials.

Existing pipelines are still bottlenecked by volumetric meshing. Segmentation-then-meshing workflows use voxel-wise segmentation~\cite{ref_3dra_segmentation2023}, marching-cubes extraction, and heavy post-processing~\cite{ref_lorensen1987,ref_updegrove2016,ref_fedele2021}; however, simulation-ready tetrahedra still rely on heuristic surface-to-volume tetrahedralization, e.g., TetGen~\cite{ref_tetgen2015,ref_tetwild2018,ref_ftetwild2020}. Such tools require case-specific tuning and yield case-dependent sampling/connectivity, limiting cross-case correspondence, population analysis~\cite{ref_virtual_chimeras2025}, and standardized learning/simulation workflows.


End-to-end image-to-mesh methods~\cite{ref_voxel2mesh2020,ref_meshdeformnet2021} reduce staging, and deformation-based models such as Voxel2Mesh, MeshDeformNet, and HeartDeformNet~\cite{ref_voxel2mesh2020,ref_meshdeformnet2021,ref_heartdeformnets2023} preserve template topology for consistent surfaces. However, two main gaps remain: they use deterministic Euclidean supervision, which inadequately models boundary uncertainty near multi-structure junctions, and they remain surface-only, requiring heuristic tetrahedralization that breaks cross-case volumetric correspondence.

\begin{figure}[!t]
\centering
\includegraphics[width=\textwidth]{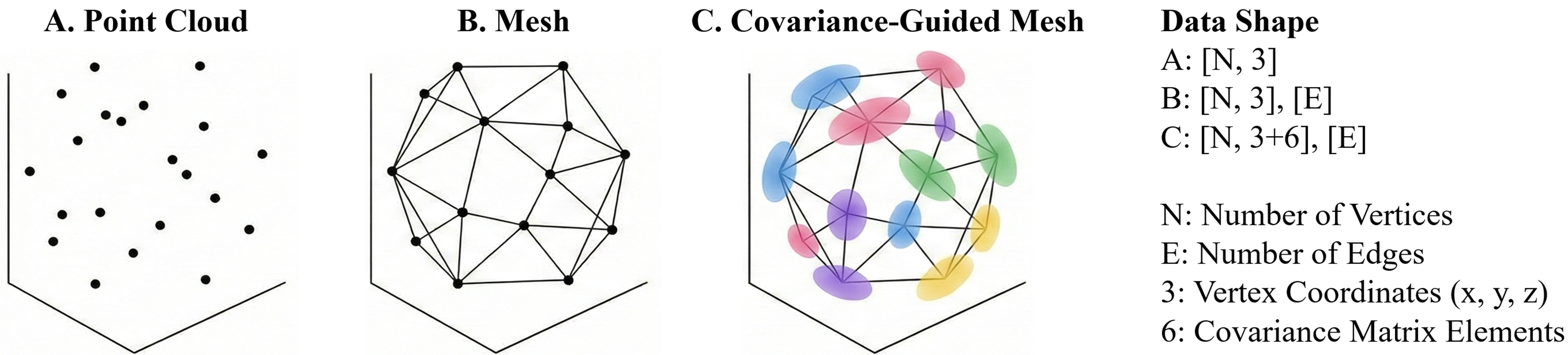}
\caption{
Schematic illustration of different representations. 
}
\label{fig:Representations}
\end{figure}

Motivated by these gaps, we seek an intermediate surface representation that captures fine details while explicitly accounting for vertex distribution, and a topology-preserving route to high-quality volumetric meshes with correspondence. We propose HeartVolMesh, a framework for cardiac digital twins with two complementary components:

\begin{enumerate}
    \item We propose a {covariance-guided graph deformation} model that represents each template vertex as an anisotropic Gaussian kernel and couples vertices through the mesh graph, enabling covariance-aware and topology-consistent surface reconstruction.
    \item We propose a {volume-to-surface registration} strategy with deformation-field propagation that warps a high-quality tetrahedral template to the patient-specific reconstructed surface, yielding tetrahedral meshes with correspondence and resolution controlled by template density.
    \item We validate in multi-structure / whole-heart settings and demonstrate improved {surface} reconstruction accuracy, as well as {new} volumetric reconstruction capability that produces tetrahedral meshes with consistent cross-case vertex correspondence, together with better volumetric mesh quality across 24 cardiac structures.
\end{enumerate}

\begin{figure}[!ht]
\centering
\includegraphics[width=\textwidth]{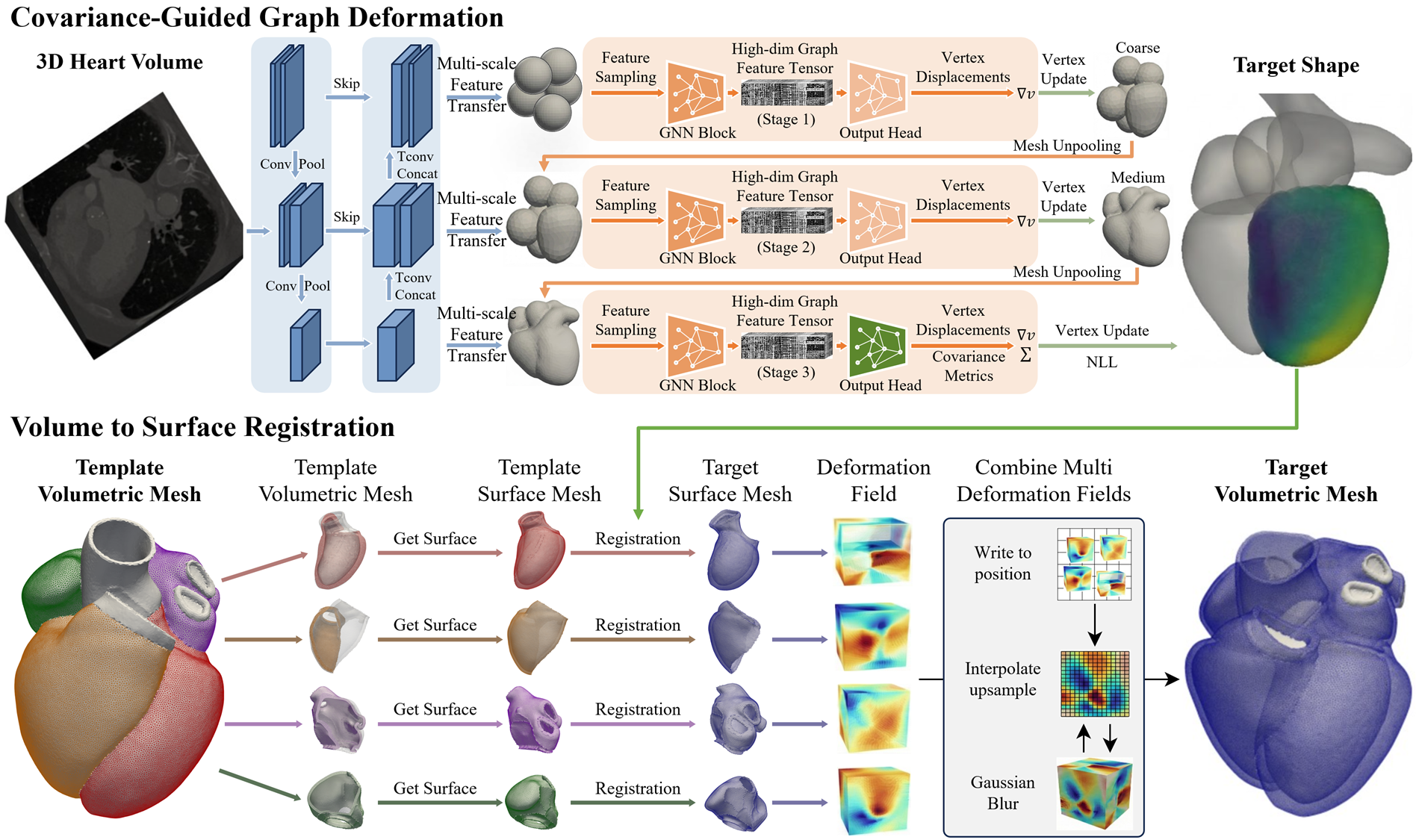}
\caption{\textbf{Overview of HeartVolMesh.}
From a CTA volume ({Input 1}), covariance-guided graph deformation reconstructs a topology-consistent target surface using anisotropic Gaussian vertices. A tetrahedral template mesh ({Input 2}) is then warped to this surface via volume-to-surface registration and deformation-field propagation, producing a patient-specific volumetric mesh with preserved connectivity and cross-case correspondence ({Output}).}

\label{fig1}
\end{figure}

\section{Methodology}

Our goal is to reconstruct topology-consistent and detail-faithful cardiac volumetric meshes from 3D medical images. The proposed {HeartVolMesh} framework consists of two complementary modules. First, we introduce {covariance-guided graph deformation} to reconstruct a topology-consistent surface while lifting each vertex to an anisotropic Gaussian kernel with a learned covariance. Second, we generate a tetrahedral mesh via a {template-driven volume-to-surface registration} strategy with deformation-field propagation, which warps a high-quality tetrahedral template to match the reconstructed surface and preserves cross-case vertex correspondence. During training, we supervise the surface module with a covariance-guided negative log-likelihood (NLL) matching loss and add lightweight mesh regularization to prevent degenerate deformations.

\subsection{Covariance-guided Graph Deformation}
\subsubsection{Theory and Representation.}
Conventional deformation-based image-to-mesh methods typically predict only vertex displacements and enforce supervision with Euclidean distances (e.g., Chamfer)~\cite{ref_pointsetgen2017}. In medical images, however, the acceptable geometric tolerance is not uniform across space or directions~\cite{ref_kendall_uncertainty2018} due to partial-volume effects, imaging noise, and thin-wall anatomy. Our {covariance-guided graph deformation} addresses this by promoting each mesh vertex from a deterministic point to an {anisotropic Gaussian kernel}. Concretely, besides predicting the deformed vertex position (Gaussian mean), we predict a full $3\times 3$ covariance matrix $\Sigma$ per vertex, which extends a classical mesh representation $[N,3],[E]$ to $[N,3{+}6],[E]$, where $E$ denotes mesh edges (graph connectivity). The covariance provides direction-dependent local geometric support, while the mesh graph couples neighboring vertices to preserve global anatomical coherence and topology consistency during deformation.

To ensure that $\Sigma$ is always symmetric positive definite (SPD) and numerically stable, we parameterize it via a Cholesky factor $\Sigma = LL^\top + \epsilon I .$
Here $L$ is a lower-triangular matrix whose diagonal is constrained to be positive (we use a softplus transform), $I$ is the identity matrix, and $\epsilon$ is a small constant that improves stability. This parameterization guarantees that $\Sigma^{-1}$ and $\log\det(\Sigma)$ are well-defined for the covariance-guided loss described in Sec.~\ref{sec:loss}.


\subsubsection{Network Architecture.}
Fig.~\ref{fig1} shows the covariance-guided graph deformation network. A 3D CNN encoder~\cite{ref_3dunet2016} extracts multi-scale volumetric features, which are interpolated at current vertex locations to obtain image-conditioned vertex features. A GNN then propagates these features over the mesh graph, using neighborhood connectivity to enforce spatial coherence and topology-consistent deformation. We use coarse-to-fine refinement: meshes are progressively unpooled and updated across stages to capture both global structure and local detail. The final head predicts vertex displacements and covariance parameters for Cholesky-based $\Sigma$ construction. In whole-heart settings, the encoder can be shared, with structure-specific mesh branches or output heads.

\subsection{Template-driven Volumetric Mesh Generation}

The bottom half of Fig.~\ref{fig1} summarizes our volume-to-surface registration pipeline. Let $\mathcal{S}$ be the template boundary surface extracted from a tetrahedral template, and $\mathcal{T}$ be the target surface reconstructed by covariance-guided graph deformation. We first perform a global similarity alignment~\cite{ref_umeyama1991} to initialize registration: $x' = sRx + t ,$ where $x\in\mathcal{S}$, $R$ is rotation, $t$ is translation, and $s$ is a scale factor. We then apply a smooth non-rigid surface registration~\cite{ref_cpd2010} to obtain per-vertex displacements on $\mathcal{S}$.

To propagate surface motion to the full volumetric template, we convert these surface displacements into a continuous 3D deformation field~\cite{ref_rueckert_ffd1999} $D(\cdot)$ defined on a voxel grid (via rasterization and interpolation/upsampling), and optionally smooth it (Gaussian blur) for regularity and stability. For multi-structure settings, we compute deformation fields per structure and combine them into a single dense field on the shared grid (write-to-position followed by interpolation and smoothing), consistent with Fig.~\ref{fig1}. Finally, we warp all vertices of the tetrahedral template: $v_T = v_S + D(v_S) ,$ where $v_S$ is a template volumetric vertex and $v_T$ is the deformed vertex. This yields a patient-specific tetrahedral mesh whose boundary matches the reconstructed surface while preserving template connectivity, correspondence, and element quality to a large extent.

\subsection{Loss Function}
\label{sec:loss}
We train the covariance-guided graph deformation module with a covariance-guided geometric matching objective plus lightweight mesh regularization. The volumetric warping module is executed at inference time as a registration-based optimization guided by the reconstructed surface.

\subsubsection{Covariance-guided NLL Matching.}
For a matched pair of points $(x,y)\in\mathbb{R}^3$ with predicted covariance $\Sigma$, we define:
\begin{equation}
\ell(x,y;\Sigma)=\tfrac{1}{2}(x-y)^\top \Sigma^{-1}(x-y) + \beta \log\det(\Sigma).
\end{equation}
The first term is a Mahalanobis distance under the learned anisotropic metric, and the second term prevents trivial covariance inflation. We use bidirectional nearest-neighbor matching (prediction-to-target and target-to-prediction) and average $\ell(\cdot)$ over matched pairs to form the surface loss $L_{\mathrm{surf}}$.

\subsubsection{Regularization and Overall Objective.}
To discourage degenerate deformations, we add standard mesh regularizers with small weights:
\begin{equation}
L = L_{\mathrm{surf}} + \lambda_{\mathrm{edge}}L_{\mathrm{edge}} + \lambda_{\mathrm{lap}}L_{\mathrm{lap}} + \lambda_{\mathrm{norm}}L_{\mathrm{norm}} .
\end{equation}
Here $L_{\mathrm{edge}}$ is edge-length regularization \cite{ref_pixel2mesh2018}, $L_{\mathrm{lap}}$ is Laplacian smoothing \cite{ref_laplacian_surface_editing2004,ref_neural_mesh_renderer2018}, and $L_{\mathrm{norm}}$ is normal consistency \cite{ref_meshrcnn2019}.

\section{Experiments and Results}
We evaluate whole-heart, multi-structure cardiac mesh reconstruction and compare with deformation-based image-to-mesh baselines. We report accuracy on both \emph{pre-warp surfaces} (network output) and \emph{post-warp volumetric meshes} (tetrahedral template warped to the predicted surface). Unless stated otherwise, results are mean~$\pm$~std over the validation set, computed per structure.

\subsection{Experimental Setup}
\label{sec:setup}

\begin{figure}[!ht]
\centering
\includegraphics[width=\textwidth]{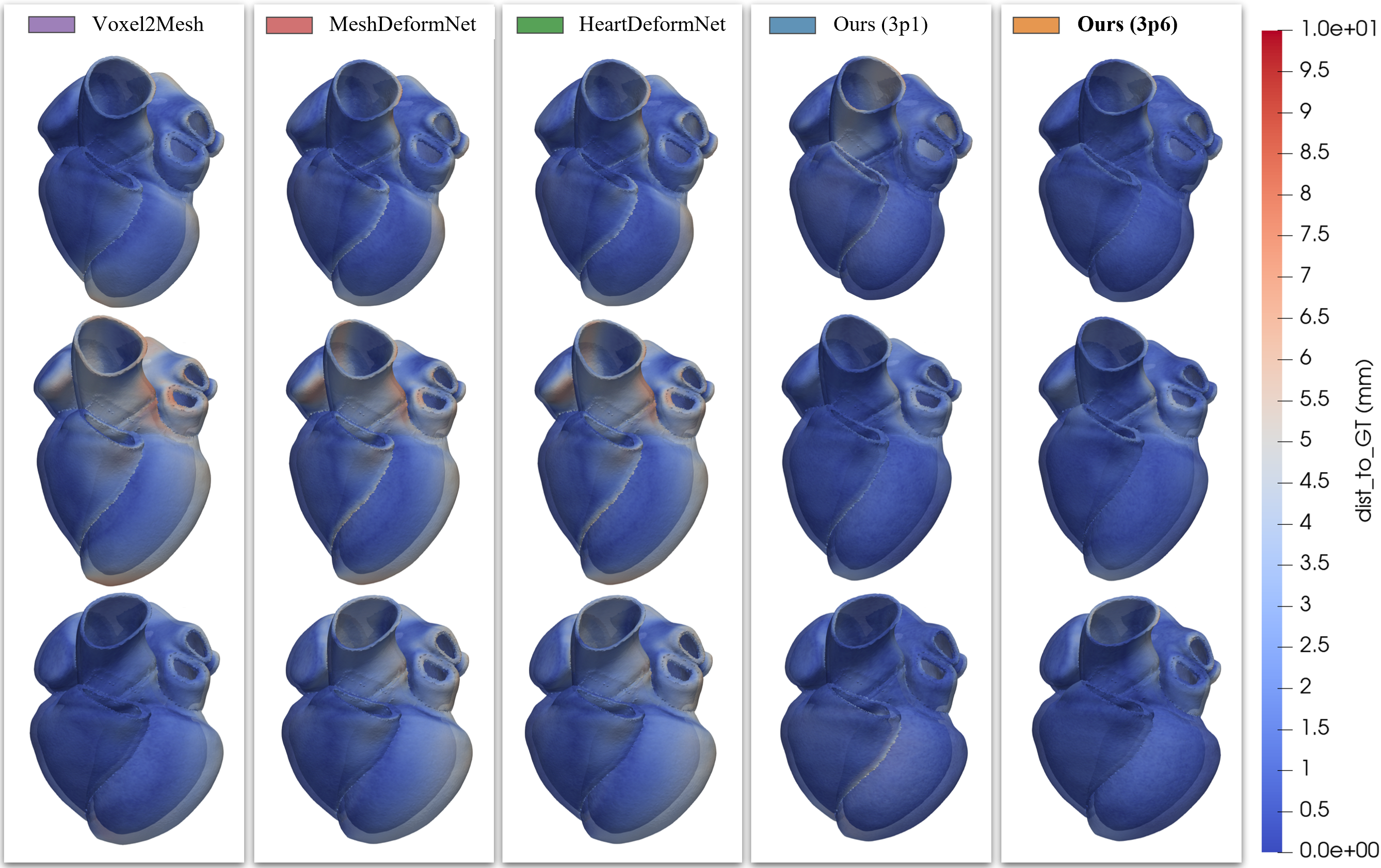}
\caption{\textbf{Qualitative volumetric reconstruction.}
Boundary surfaces extracted from the final tetrahedral meshes are visualized for LA/LV/RA/RV/Myo. {Ours3p6} produces sharper boundaries and fewer outliers, especially near inter-structure junctions.}
\label{fig:result2}
\end{figure}

We used a multi-centre in-house dataset of 900 patients and 4,000 temporal instances, split at the patient level into 800/100 patients for training/validation (3,600/400 instances). Reference pseudo-GT meshes were obtained from TotalSegmentator outputs with manual refinement, surface extraction, and the same non-learned template-warping procedure as in Sec.~2.2.
All models were implemented in PyTorch and trained on one NVIDIA A100 80GB GPU. The 3D CNN encoder and mesh GNN were jointly optimized with Adam for 100 epochs, batch size 1, using $128^3$ crops. The deformation field resolution was $300^3$. We used $\beta=0.5$ and $\lambda_{edge}=\lambda_{lap}=\lambda_{norm}=0.1$. Template-driven volumetric warping was applied after surface prediction and was not jointly trained.
We evaluated surface reconstruction on LA, LV, RA, RV, and LVMyo, and volumetric reconstruction on LAMyo, LVMyo, RAMyo, and RVMyo using boundary surfaces extracted from the warped tetrahedral meshes. Metrics included CD, HD95, NC, inverted elements, minimum scaled Jacobian, and minimum dihedral angle. We compared Voxel2Mesh, MeshDeformNet, HeartDeformNet, Ours3p1, and Ours3p6, applying the same warping pipeline to all methods for fair volumetric evaluation.

\subsection{Comparison with State-of-the-Art Methods}

\textbf{Volumetric meshes.}
Table~\ref{tab:vol_surf_merged_split} (left) evaluates the {boundary surfaces} extracted from the final warped tetrahedral meshes, measuring how accurately the resulting volumetric meshes align to the target anatomy. Our {Ours3p6} achieves the best CD/HD95 and the highest NC across all four volumetric targets, demonstrating consistent improvements on both the thin-wall atrial myocardium (LAMyo) and the highly curved ventricular regions (LVMyo/RVMyo). While {Ours3p1} already outperforms prior work, {Ours3p6} further benefits from anisotropic covariance modelling and covariance-guided supervision, leading to more accurate and better-aligned volumetric boundaries.
Beyond boundary accuracy, we also report tetrahedral element-quality statistics (Sec.~\ref{sec:setup}) to assess simulation readiness. To ensure a fair comparison, we plug all methods into the same warping module (our proposed warping pipeline). Across 400 test instances (each with $\sim$3M tetrahedra), ours {HeartVolMesh} yields {0.0\%} inverted elements, with a minimum scaled Jacobian of $0.0416 \pm 0.0152$ and a minimum dihedral angle of $5.56^{\circ} \pm 1.41^{\circ}$.

\textbf{Surface meshes.}
Table~\ref{tab:vol_surf_merged_split} (right) reports surface-mesh accuracy {before warping}. Our {Ours3p6} achieves the best CD/HD95 on all five surfaces and consistently strong NC, indicating improved localization under boundary ambiguity. The performance gap between {Ours3p6} and {Ours3p1} further supports the benefit of anisotropic covariance modelling for surface learning.

\begin{table*}[t]
\caption{Left: volumetric meshes (boundary surfaces). Right: surface meshes. Metrics: CD/HD95 in mm (lower is better) and NC in \% (higher is better). Mean~$\pm$~std. Bold indicates statistically significant improvement (t-test $p<0.05$).}
\label{tab:vol_surf_merged_split}
\centering
\setlength{\tabcolsep}{0.2pt}
\scriptsize
\resizebox{\textwidth}{!}{
\begin{tabular}{|l|cccc|ccccc|}
\hline
\textbf{Category}& \multicolumn{4}{c|}{\textbf{Volumetric mesh (boundary surfaces)}} & \multicolumn{5}{c|}{\textbf{Surface mesh}} \\
\hline

\multicolumn{10}{c}{\textbf{CD} $\downarrow$} \\
\hline
\textbf{Method} & \textbf{LVMyo} & \textbf{RVMyo} & \textbf{LAMyo} & \textbf{RAMyo} & \textbf{LA} & \textbf{LV} & \textbf{Myo} & \textbf{RA} & \textbf{RV} \\
\hline
Voxel2Mesh            & 5.9$\pm$0.5 & 4.3$\pm$0.7 & 4.6$\pm$1.5 & 3.7$\pm$0.9 & 2.7$\pm$0.2 & 3.5$\pm$0.3 & 4.9$\pm$0.4 & 3.5$\pm$0.3 & 4.1$\pm$0.5 \\
MeshDeformNet         & 5.8$\pm$0.5 & 4.4$\pm$0.5 & 4.2$\pm$1.4 & 4.3$\pm$0.7 & 2.8$\pm$0.2 & 3.7$\pm$0.4 & 5.4$\pm$0.5 & 3.9$\pm$0.4 & 4.2$\pm$0.8 \\
HeartDeformNet        & 5.8$\pm$0.5 & 4.5$\pm$0.6 & 4.3$\pm$1.6 & 4.3$\pm$0.8 & 2.8$\pm$0.2 & 3.7$\pm$0.4 & 5.2$\pm$0.5 & 3.7$\pm$0.5 & 4.1$\pm$0.6 \\
Ours 3p1              & 4.5$\pm$1.0 & 3.9$\pm$0.9 & 4.0$\pm$1.2 & 3.6$\pm$0.9 & 2.6$\pm$0.2 & 3.0$\pm$0.4 & 4.1$\pm$0.4 & 3.4$\pm$0.3 & 3.8$\pm$0.4 \\
\textbf{Ours 3p6}      & \textbf{3.6}$\pm$\textbf{0.4} & \textbf{3.1}$\pm$\textbf{0.4} & {3.0}$\pm${0.3} & {3.0}$\pm${0.4} &
\textbf{2.5}$\pm$\textbf{0.2} & \textbf{3.0}$\pm$\textbf{0.4} & \textbf{4.0}$\pm$\textbf{0.4} & \textbf{2.9}$\pm$\textbf{0.3} & \textbf{3.3}$\pm$\textbf{0.4} \\
Ablation-A            & 5.8$\pm$0.5 & 4.5$\pm$0.6 & 4.3$\pm$1.6 & 4.3$\pm$0.8 & 2.8$\pm$0.2 & 3.9$\pm$0.5 & 5.3$\pm$0.6 & 3.8$\pm$0.6 & 4.5$\pm$0.9 \\
Ablation-B            & 3.9$\pm$0.5 & 3.4$\pm$0.5 & 2.9$\pm$0.4 & 3.0$\pm$0.3 & / & / & / & / & / \\
Ablation-C            & 5.1$\pm$0.9 & 4.5$\pm$0.6 & 6.6$\pm$1.4 & 6.4$\pm$1.4 & 3.1$\pm$0.2 & 3.4$\pm$0.4 & 4.3$\pm$0.4 & 3.3$\pm$0.4 & 4.0$\pm$0.4 \\
\hline

\multicolumn{10}{c}{\textbf{HD95} $\downarrow$} \\
\hline
\textbf{Method} & \textbf{LVMyo} & \textbf{RVMyo} & \textbf{LAMyo} & \textbf{RAMyo} & \textbf{LA} & \textbf{LV} & \textbf{Myo} & \textbf{RA} & \textbf{RV} \\
\hline
Voxel2Mesh            & 5.8$\pm$0.7 & 4.8$\pm$1.2 & 5.2$\pm$1.7 & 4.2$\pm$1.3 & 2.4$\pm$0.2 & 3.3$\pm$0.5 & 4.8$\pm$0.4 & 3.8$\pm$0.7 & 3.9$\pm$0.7 \\
MeshDeformNet         & 5.4$\pm$0.6 & 4.9$\pm$1.0 & 4.7$\pm$1.5 & 4.7$\pm$1.0 & 2.5$\pm$0.2 & 3.6$\pm$0.5 & 5.4$\pm$0.6 & 4.5$\pm$0.8 & 4.6$\pm$2.0 \\
HeartDeformNet        & 5.4$\pm$0.7 & 4.9$\pm$1.1 & 4.8$\pm$1.7 & 4.7$\pm$1.1 & 2.5$\pm$0.2 & 3.8$\pm$0.8 & 5.4$\pm$0.7 & 4.2$\pm$1.2 & 4.5$\pm$1.6 \\
Ours 3p1              & 4.2$\pm$1.2 & 3.8$\pm$1.3 & 4.4$\pm$1.5 & 3.9$\pm$1.3 & 2.3$\pm$0.1 & 2.8$\pm$0.4 & 3.8$\pm$0.4 & 3.7$\pm$1.0 & 3.7$\pm$0.8 \\
\textbf{Ours 3p6}      & \textbf{3.2}$\pm$\textbf{0.4} & \textbf{2.8}$\pm$\textbf{0.3} & {3.0}$\pm${0.5} & \textbf{2.9}$\pm$\textbf{0.6} &
\textbf{2.2}$\pm$\textbf{0.2} & \textbf{2.8}$\pm$\textbf{0.5} & \textbf{3.6}$\pm$\textbf{0.3} & \textbf{2.8}$\pm$\textbf{0.5} & \textbf{3.1}$\pm$\textbf{0.5} \\
Ablation-A            & 5.4$\pm$0.7 & 5.0$\pm$1.1 & 4.8$\pm$1.7 & 4.8$\pm$1.1 & 2.6$\pm$0.3 & 3.8$\pm$0.7 & 5.8$\pm$0.7 & 4.4$\pm$1.2 & 5.1$\pm$2.3 \\
Ablation-B            & 3.7$\pm$0.4 & 3.1$\pm$0.5 & 2.8$\pm$0.4 & 3.0$\pm$0.5& / & / & / & / & / \\
Ablation-C            & 5.0$\pm$1.1 & 5.2$\pm$1.0 & 7.7$\pm$1.9 & 7.5$\pm$1.4 & 2.8$\pm$0.1 & 3.2$\pm$0.5 & 4.0$\pm$0.4 & 3.5$\pm$0.8 & 4.0$\pm$0.8 \\
\hline

\multicolumn{10}{c}{\textbf{Normal Consistency} (\%) $\uparrow$} \\
\hline
\textbf{Method} & \textbf{LVMyo} & \textbf{RVMyo} & \textbf{LAMyo} & \textbf{RAMyo} & \textbf{LA} & \textbf{LV} & \textbf{Myo} & \textbf{RA} & \textbf{RV} \\
\hline
Voxel2Mesh            & 94.1$\pm$0.7 & 91.5$\pm$1.7 & 83.6$\pm$3.9 & 89.9$\pm$3.2 & 89.3$\pm$2.9 & 87.8$\pm$2.7 & 86.4$\pm$3.4 & 87.7$\pm$2.0 & 87.3$\pm$2.9 \\
MeshDeformNet         & 93.9$\pm$0.7 & 90.6$\pm$1.2 & 83.4$\pm$4.0 & 87.6$\pm$4.1 & 89.2$\pm$2.9 & 87.6$\pm$3.0 & 86.1$\pm$3.0 & 87.5$\pm$1.8 & 86.9$\pm$2.4 \\
HeartDeformNet        & 93.9$\pm$0.7 & 90.4$\pm$1.4 & 83.1$\pm$4.2 & 87.5$\pm$4.2 & 89.1$\pm$3.1 & 87.6$\pm$3.0 & 86.1$\pm$3.3 & 87.5$\pm$2.1 & 87.2$\pm$2.8 \\
Ours 3p1              & 94.3$\pm$1.1 & 92.1$\pm$2.3 & 84.6$\pm$4.6 & 90.1$\pm$5.2 & 89.2$\pm$2.8 & 88.2$\pm$3.1 & 87.0$\pm$3.3 & 87.5$\pm$2.4 & 86.3$\pm$3.3 \\
\textbf{Ours 3p6}      & \textbf{95.3}$\pm$\textbf{0.7} & \textbf{94.2}$\pm$\textbf{0.9} & {87.7}$\pm${2.5} & \textbf{92.2}$\pm$\textbf{3.8} &
\textbf{89.5}$\pm$\textbf{2.6} & \textbf{88.7}$\pm$\textbf{2.8} & \textbf{87.6}$\pm$\textbf{3.0} & \textbf{88.3}$\pm$\textbf{2.3} & \textbf{87.3}$\pm$\textbf{3.3} \\
Ablation-A            & 93.8$\pm$0.8 & 90.5$\pm$1.4 & 83.2$\pm$4.2 & 87.6$\pm$4.1 & 89.0$\pm$3.3 & 87.3$\pm$3.1 & 86.0$\pm$3.3 & 87.3$\pm$2.1 & 86.3$\pm$3.1 \\
Ablation-B            & 95.1$\pm$0.7 & 93.4$\pm$1.0 & 87.7$\pm$1.4 & 91.9$\pm$2.6 & / & / & / & / & / \\
Ablation-C            & 93.1$\pm$1.2 & 90.3$\pm$1.5 & 78.5$\pm$4.7 & 84.0$\pm$4.8 & 87.9$\pm$3.1 & 87.2$\pm$2.9 & 86.8$\pm$2.9 & 86.7$\pm$2.9 & 84.7$\pm$3.6 \\
\hline
\end{tabular}
}
\end{table*}


\begin{figure}[!ht]
\centering
\includegraphics[width=\textwidth]{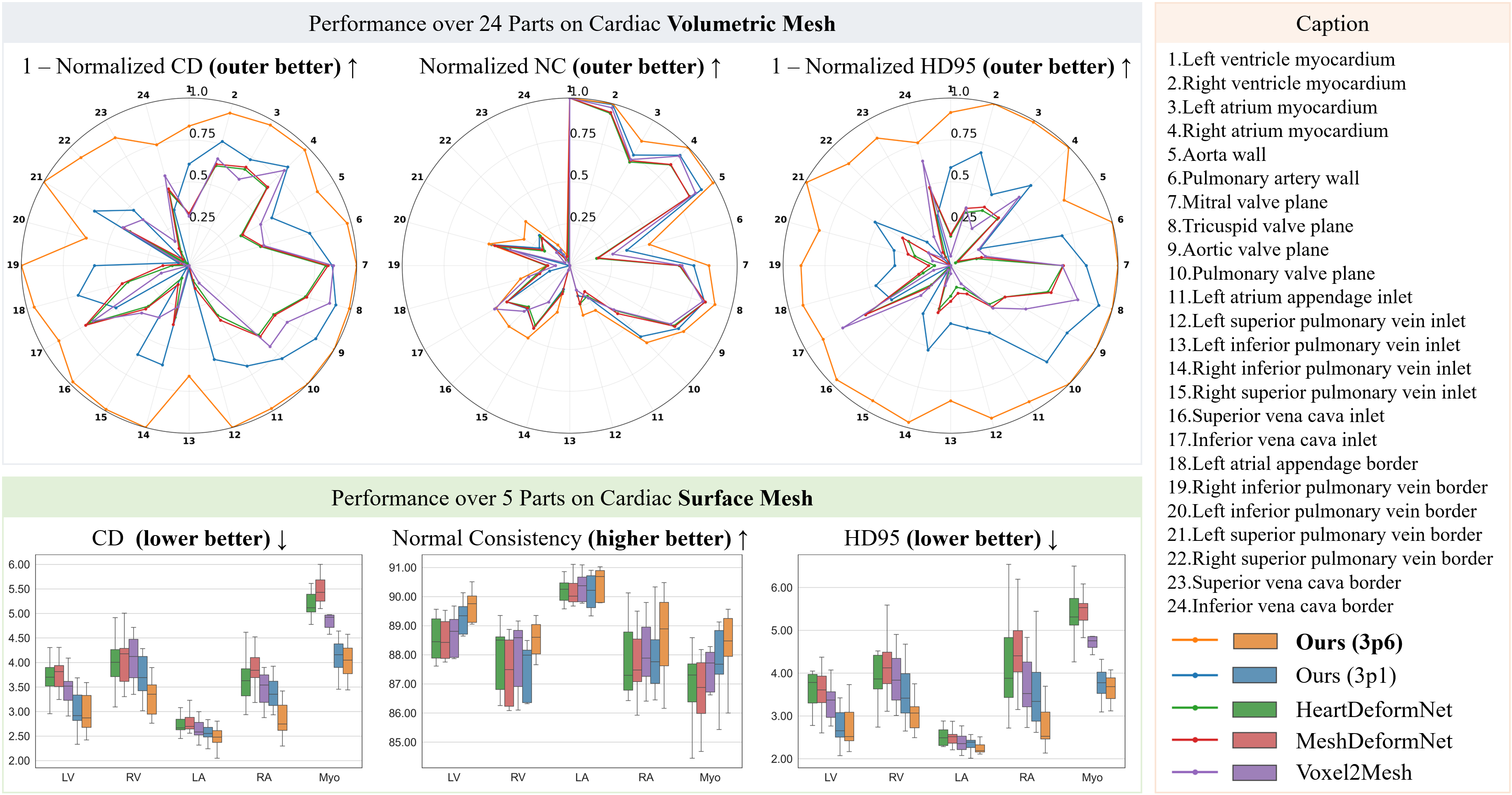}
\caption{\textbf{Quantitative summary on multi-centre in house data.}
Top: normalized radar plots over 24 volumetric parts. Bottom: boxplots over five surface structures. {Ours3p6} shows lower errors and reduced variance.}
\label{fig:result}
\end{figure}

\subsection{Ablation Study}
Ablation-A replaces the covariance-guided NLL with Chamfer-only supervision. Ablation-B (same surface mesh result with Ours3p6) uses a coarser deformation-field for volumetric warping with reduced motion-coherence strength. Ablation-C removes mesh regularization (edge/Laplacian/normal) during surface learning.
Ablation-C degrades both volumetric and surface performance, indicating that mesh regularization is important to avoid degenerate deformations that propagate to tetrahedral meshes. Ablation-A underperforms the full model, suggesting the benefit of covariance-guided supervision. Ablation-B shows mixed results, implying that overly coarse warping can oversmooth thin-wall regions.


\section{Conclusion}

We presented HeartVolMesh for cardiac volumetric mesh reconstruction with cross-case vertex correspondence from 3D images via covariance-guided graph deformation and template-conditioned warping. On multi-centre in-house data, our method consistently improves boundary accuracy. 
We trade end-to-end meshing (which typically yields a single fixed mesh specification after training) for a template-conditioned design that supports diverse simulation mesh specifications (e.g., density and element type).
Future work will extend to cine CMR, benchmark multiple template specifications, and validate simulations.

\begingroup
\emergencystretch=2em
\subsubsection{Acknowledgements.}
AFF acknowledges support from the Royal Academy of Engineering under the RAEng Chair in Emerging Technologies
(INSILEX CiET1919 19), ERC Advanced Grant--UKRI Frontier Research Guarantee
(INSILICO EP Y030494 1), the UK Centre of Excellence on in-silico Regulatory Science and Innovation
(UK CEiRSI) (10139527), the National Institute for Health and Care Research (NIHR) Manchester Biomedical Research Centre
(BRC) (NIHR203308), the BHF Manchester Centre of Research Excellence
(RE 24 30017), and the CRUK RadNet Manchester
(C1994 A28701). AZ/FL/AFF acknowledges support from NVIDIA Academic Grant Program.
\par
\endgroup

\subsubsection{Disclosure of Interests.}
AF is Co-founder and shareholder of OculomeX Health Ltd, adsilico Ltd, and Synaptive Consulting Ltd.


\end{document}